\documentclass[journal]{IEEEtran}
\usepackage{amsmath}

\usepackage{lastpage,fancyhdr,graphicx}
\graphicspath{ {./svg_figures/} }

\usepackage[hyphens]{url}
\usepackage{array}

\usepackage{hyperref}
\newcolumntype{+}{!{\vrule width 2pt}}

\newlength\savedwidth

\newcommand\thickhline{\noalign{\global\savedwidth\arrayrulewidth\global\arrayrulewidth 2pt}%
	\hline
	\noalign{\global\arrayrulewidth\savedwidth}}

\newcommand\eatpunct[1]{}

\ifCLASSINFOpdf
\else
\fi
\begin{document}
\title{Focal onset seizure prediction using convolutional networks}

\author{Haidar~Khan*,
        Lara~Marcuse,
        Madeline~Fields,
        Kalina~Swann,
        and B\"{u}lent~Yener,~\IEEEmembership{Fellow,~IEEE}% <-this % stops a space
\thanks{Manuscript received May 17, 2017; revised November 8, 2017. This
	work was supported in part by the U.S. National Science Foundation under
	Grant 1302231.}% <-this % stops a space
\thanks{L. Marcuse and M. Fields are with The Mount Sinai Epilepsy Center, Mount Sinai Hospital, 
	New York, NY, 10029 USA. H. Khan, K. Swann, and B. Yener are with the Department
	of Computer Science, Rensselaer Polytechnic Institute, Troy,
	NY, 12180 USA (correspondence e-mail: khanh2@rpi.edu).}
\thanks{Copyright (c) 2016 IEEE. Personal use of this material is permitted. 
	However, permission to use this material for any other purposes must be obtained 
	from the IEEE by sending an email to pubs-permissions@ieee.org.}}% <-this % stops a space

\maketitle

\thispagestyle{fancy}
\begin{abstract}
	\textit{ Objective:}	This work investigates the hypothesis that focal seizures can be predicted using scalp electroencephalogram (EEG) data. Our first aim is to learn features that distinguish between the interictal and preictal regions. The second aim is to define a prediction horizon in which the prediction is as accurate and as early as possible, clearly two competing objectives.
	\textit{Methods:} Convolutional filters on the wavelet transformation of the EEG signal are used to define and learn quantitative signatures for each period: interictal, preictal, and ictal. The optimal seizure prediction horizon is also learned from the data as opposed to making an a priori assumption.
	\textit{Results:} Computational solutions to the optimization problem indicate a ten-minute seizure prediction horizon. This result is verified by measuring Kullback-Leibler divergence on the distributions of the automatically extracted features.
	\textit{Conclusion:} The results on the EEG database of 204 recordings demonstrate that (i) the preictal phase transition occurs approximately ten minutes before seizure onset, and (ii) the prediction results on the test set are promising, with a sensitivity of 87.8\% and a low false prediction rate of 0.142 FP/h. Our results significantly outperform a random predictor and other seizure prediction algorithms.
	\textit{Significance:} We demonstrate that a robust set of features can be learned from scalp EEG that characterize the preictal state of focal seizures.    

\end{abstract}

\begin{IEEEkeywords}
automatic feature extraction, convolutional neural networks, deep learning, focal seizures, preictal period, scalp EEG, seizure prediction  
\end{IEEEkeywords}

% \fi
\IEEEpeerreviewmaketitle

\section{Introduction}
%\IEEEPARstart{T}{his} demo file is intended to serve as a ``starter file''
%for IEEE journal papers produced under \LaTeX\ using
%IEEEtran.cls version 1.8 and later.
\IEEEPARstart{W}{orldwide}, there are approximately 65 million people with epilepsy, more than Parkinson’s disease, Alzheimer’s disease, and multiple sclerosis combined. Epileptic seizures are unpredictable, occurring often without warning.  This contributes to the anxiety, morbidity, and mortality of the illness. The seizure prediction problem has, until recently, evaded success from computational approaches utilizing electroencephalogram (EEG) data. The difficulty of the problem arises from the lack of a general and specifiable definition of the phase transition between interictal and preictal periods of the EEG signal.
\paragraph*{\eatpunct}  
The shift from the hand-crafted design of features for machine learning systems to the merging of feature extraction with the learning process has proved successful on many interesting tasks, ranging from handwritten digit recognition to language translation. The idea of applying automatic feature extraction techniques to new data types other than images or natural language is a promising one. One example is the application of deep learning and automatic feature extraction to genomics data, yielding novel insights into patterns in DNA sequences~\cite{Shrikumar2016}.
\paragraph*{\eatpunct}
In this work we apply automatic feature extraction techniques to predict seizures from scalp EEG, towards constructing a system to alert patients about oncoming seizures.
\paragraph*{\eatpunct}
This paper is organized as follows. The rest of Section I covers the relevant prior work. Section II details the methods used in this study. Section III presents the discovered preictal phase transition and the results on the test set. Section IV compares this work with other seizure prediction methods. Finally, we discuss the results in Section V and conclude in Section VI.

\subsection{Related Work}
\paragraph*{\eatpunct}
In the last two decades, research in the area of seizure prediction has matured as a result of the formalization of the problem and the availability of EEG data~\cite{Gadhoumi2016a}. The underlying assumption of seizure prediction is that a difference exists in the brain waves between the interictal and preictal states. Many previous methods have failed to reliably predict seizures~\cite{Harrison2005,Freestone2015}. However, the algorithms of multiple contestants in recent seizure prediction competitions on Kaggle~\cite{Brinkmann2016} functioned at above random levels at accurately classifying interictal vs. preictal data, demonstrating the feasibility of seizure prediction. An implanted device~\cite{Cook2013b} was the first study demonstrating prospective seizure prediction on long-term intracranial EEG collected in an ambulatory setting.
\paragraph*{\eatpunct}
Work on the seizure prediction problem has been conducted on two sources of EEG data, intracranial EEG and scalp EEG. Due to the popularity of the open Freiburg EEG dataset~\cite{freiburg2010data} (now combined with EPILEPSIAE~\cite{Ihle2012}), much of the early work on the seizure prediction problem utilized the available intracranial EEG data to develop algorithms for an implantable seizure prediction device. With the compilation of the EPILEPSIAE database, which contains a growing number of scalp EEGs, attention is being shifted towards the possibility of an external seizure prediction device.
\paragraph*{\eatpunct}
The nature of data collected by intracranial EEG and scalp EEG differs greatly. Scalp EEG is readily available and is not invasive. However, it is more prone to artifacts introduced by shifting electrodes, muscle interference, and the effects of volume conduction. Intracranial EEG has a better signal-to-noise ratio than scalp EEG and can target specific areas of the brain directly. While most work has focused directly on human EEG, some studies have used canine intracranial EEG from dogs with naturally occurring epilepsy to explore the seizure prediction problem~\cite{Brinkmann2016, Cherkassky2015,Korshunova2017}.
\paragraph*{\eatpunct}
Seizure prediction systems using intracranial or scalp EEG signals rely on moving window analysis on extracted features to generate predictions. One of the main challenges for accurate prediction is extracting and evaluating linear and nonlinear univariate and bivariate features from the signal. Seizure prediction methods have reported encouraging results using extracted linear features from the EEG signal. Autoregressive coefficients~\cite{Chisci2010}, spike rate~\cite{Karoly2016,Li2013} Hjorth parameters, spectral band power, and cross correlation are some linear features considered~\cite{Mormann2007,Gadhoumi2016a,Nagaraj2015}. The advent of the theory of dynamical systems introduced a number of non-linear features, such as the dynamical similarity index~\cite{LeVanQuyen1999}, largest Lyapunov exponent ~\cite{Iasemidis2003}, phase synchronization~\cite{Cho2017a}, attractor states~\cite{Chu2017}, and combinations of non-linear features~\cite{Aarabi2012}. Other extracted features include diffusion kernels~\cite{Duncan2013} and synchronization graphs~\cite{Dhulekar2014}.
\paragraph*{\eatpunct} 
Once a system extracts a set of features, two main approaches have been taken for prediction; statistical and algorithmic. Statistical methods, such as multivariate time series analysis techniques~\cite{Schelter2006}, attempt to estimate the usefulness of extracted features to the prediction problem retrospectively. Algorithmic methods use combinations of features or model parameters, such as~\cite{Aarabi2014,Aarabi2017}, and an explicit threshold to predict seizures prospectively~\cite{LeVanQuyen2005}. Support vector machines~\cite{Mirowski2009}, random forests~\cite{Brinkmann2016}, autoencoders~\cite{Stober2015} and other machine learning algorithms combined with feature extraction and selection methods have been applied to the seizure prediction problem. Formulating the problem as an instance of anomaly detection is also proposed, but due to the complexity of the underlying system, it is difficult to establish a robust ‘baseline’ model that does not result in an overwhelming number of false positives, which is highly undesirable in this context~\cite{Simard2003}. The closest work to this one applies convolutional neural networks (CNN) to selected frequency bands on intracranial EEG~\cite{Mirowski2008a, Mirowski2009}. They report improved results using CNNs over support vector machines on extracted features, but do not use CNN to extract features directly from the EEG signal.
\paragraph*{\eatpunct}
The length of the preictal period, the period before the seizure occurs in which it is possible to anticipate the seizure, has often been left as a design choice and has varied from hours to minutes~\cite{Mormann2007}. Some approaches to selecting the appropriate prediction horizon have been proposed, such as~\cite{Bandarabadi2015}, but suffer from being dependent on a specific group of features. Work by~\cite{Cook2013b,Freestone2017} show that seizures captured by intracranial EEG recordings are patient-specific, but it is not clear if this affects the length of the preictal period or applies to seizures captured by scalp EEG data.  Methods to statistically validate results based on the length of the preictal period have been developed and accepted, including time-series surrogates and comparison to an unspecific random predictor~\cite{Schelter2006}. 

\section{Methods}
\subsection{Problem Definition}
\paragraph*{\eatpunct}
Underlying the search for the prediction horizon is the assumption that changes in the brain occur prior to seizure onset that make the seizure nearly inevitable. Despite this, seizure prediction based on EEG data has posed a challenge to the research community due to the absence of a clear and robust definition of the problem~\cite{Osorio2011}. The seizure prediction problem is defined as anticipating a seizure within some prediction horizon, or time window before seizure onset. Defining concretely the prediction horizon is difficult, since the optimal time window for prediction is not well understood. Our goal is to derive a justifiable patient-independent prediction horizon (preictal period) directly from the data while searching for early EEG predictors of the phase transition between the interictal and preictal periods. For the purpose of this study, we assume that a preictal phase exists for all focal seizures and that there is an inflection point between interictal and preictal states.
\subsection{Datasets and Preprocessing}
\paragraph*{\eatpunct}
In this study, we trained our model and evaluated its performance on two independent datasets: (i) data collected from The Mount Sinai Epilepsy Center at the Mount Sinai Hospital (MSSM), and (ii)  a subset of the public CHB-MIT EEG database (CHB-MIT)~\cite{Goldberger2000, Shoeb2009a}. The datasets are composed of two types of recordings, interictal recordings without seizures and recordings with seizures. We use recordings with seizures to learn relevant features, determine the preictal period, and evaluate the sensitivity of the model. Interictal recordings are important to test the specificity of the model and estimate the false prediction rate. The system was trained and crossvalidated on 96 EEG recordings with seizures and 20 EEG recordings without seizures. Testing was done on 35 EEG recordings with seizures and 53 EEG recordings without seizures. In total, this study analyzed 131 recordings with seizures and 73 recordings without seizures. The recordings were collected using the standard 10-20 system for electrode placement with a bipolar montage. As a data preparation step, the EEG was filtered with a 128Hz low-pass filter and verified to have nonlinear structure using the BDS nonlinearity test with the appropriate ARIMA model~\cite{Broock1996}.
\subsubsection{MSSM EEG Dataset}
\paragraph*{\eatpunct}
The MSSM dataset contained 86 scalp EEG recordings from 28 patients with epilepsy. The recordings were made from continuous EEG studies that utilized XLTEK equipment with 22 inputs and a sampling rate of 256Hz. The duration of the continuous EEG study varied from 2-8 days. All patient information was de-identified and the relevant EEG activity was converted into raw data using the European Data Share format~\cite{kemp2003}.
\paragraph*{\eatpunct}
The seizures were all focal with variable seizure onset zones, primarily temporal and frontal. All EEG data was selected, reviewed, and de-identified by two electroencephalographers. EEGs with electrode artifacts affecting more than one electrode were excluded.  However, there was no attempt to remove sleep-wake transitions, eye blinks, movement, and chewing artifacts. Each subject was assigned a number from 1-28 without any identifying information. Sixty-one of the 86 recordings contained a seizure with onset times marked, and the remaining are interictal data only. The electroencephalographers were unable to distinguish any preictal signal on the raw EEG in any of the recordings. The interictal recordings were typically 60 minutes in length.  Prior to a seizure, 75 minutes of EEG data was included.
\paragraph*{\eatpunct}
The types of seizures captured include subclinical as well as clinical focal seizures.  No seizures with generalized onsets were used.  Each patient in this study had more than one seizure during their recording.
\subsubsection{CHB-MIT EEG Dataset}
\paragraph*{\eatpunct}
The CHB-MIT database contains scalp EEG data collected from 22 patients with 9-42 recordings for each patient, originally collected for the seizure detection problem~\cite{Shoeb2010}. 136 of the recordings contain one or more labeled seizure and 509 of the recordings contain no seizure activity. The recordings were collected at a sampling rate of 256Hz. We selected a subset of recordings suitable for our study from this dataset for training and testing. Specifically, we selected 68 of the 129 seizure recordings which met the following two requirements: (a) contained only one seizure event and (b) contained at least 30 minutes of EEG data before the seizure event. We also randomly selected 50 recordings with no seizure activity from the CHBMIT database to measure the specificity of the model.
\subsubsection{From Raw Data to Wavelet Tensors}

\begin{figure}[!h]
	\includegraphics[width=\linewidth]{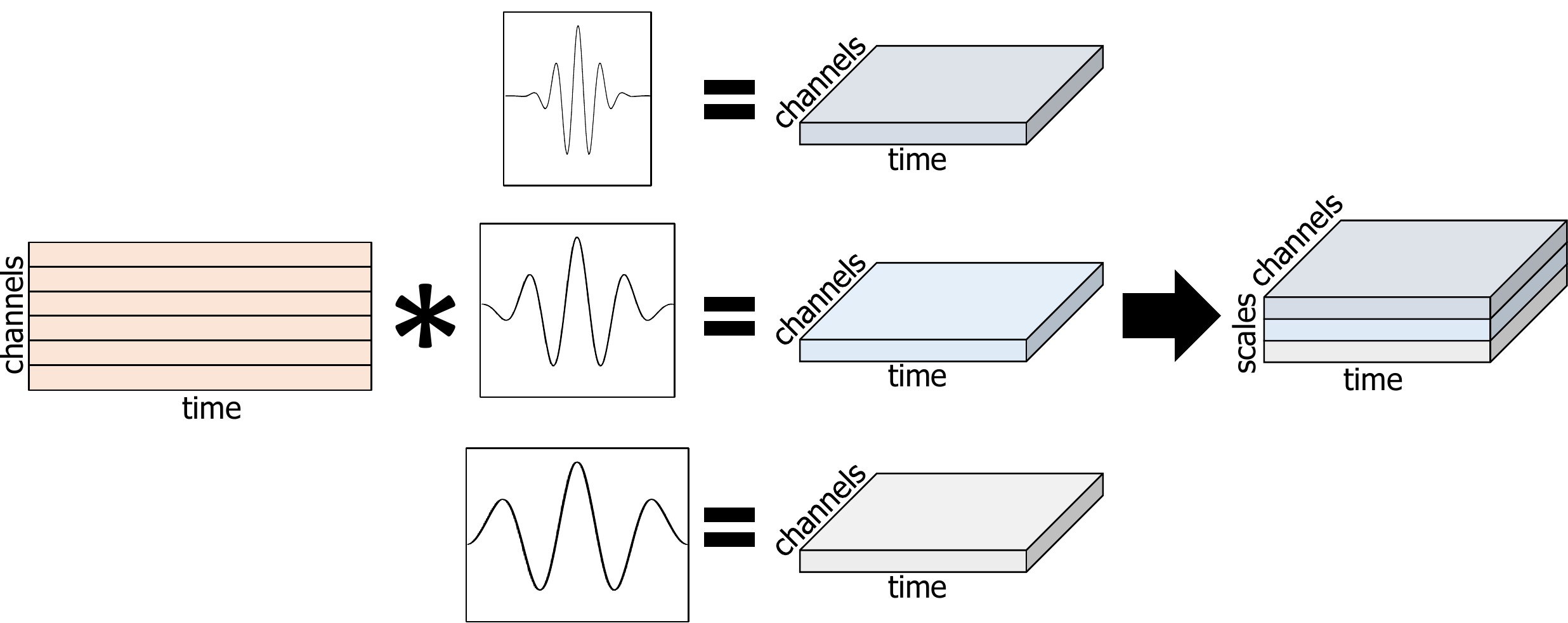}
	\caption{{\bf EEG wavelet tensor.}
		Wavelet tensor constructed by convolving each channel of the EEG signal with wavelet functions at different scales. There are 22 channels and 10 scales in each wavelet tensor.}
	\label{wavelet_tensor}
\end{figure}

\paragraph*{\eatpunct}
The wavelet transform on an EEG signal transforms the signal from the time view to a combination of the time and frequency view, which is useful both in clinical and computational analysis~\cite{Kim2000}. Applying the continuous wavelet transform to each channel of the EEG yields a tensor of wavelet coefficients with three modes; time, scales, and channels as shown in Fig~\ref{wavelet_tensor}.

\paragraph*{\eatpunct}
We use the wavelet transform to provide a window into the frequency domain to aid in our analysis and as input to the convolutional neural network (CNN). This was motivated by previous success with the use of tensor decomposition and analysis on wavelet tensors for the seizure detection problem~\cite{Acar2007} and  with a similar approach to genomic data~\cite{Shrikumar2016}. Other benefits of using this transformed signal include revealing multiscale frequency information at each time point and isolating noise~\cite{Daubechies1990}. In theory, all of these benefits could be achieved by several convolutional layers trained on the raw EEG signal~\cite{Krizhevsky2012} but at the cost of needing more training data and training time. Training on the wavelet-transformed signal achieves a “deeper” network without the time and data penalty of training an additional set of convolutional layers. In addition, the training algorithm for the CNN is extended to include the scale parameters of the wavelet transform.
\paragraph*{\eatpunct}

\subsection{Methodology: Overview and Background}
\paragraph*{\eatpunct}
The first aim of our approach is to extract features from the EEG signal that can be used to distinguish between different functional states. We use CNN to extract features from the EEG signal and focus on explicitly differentiating between preictal, ictal, and interictal examples. Since the true preictal period length is unknown, the second aim is to estimate the length of the preictal period and the optimal prediction horizon. 
%As shown in Fig~\ref{preictal}, 
As there are many possible candidate preictal period lengths, we use cross-validation to choose the appropriate length.

\subsubsection{Aim 1: Convolutional Neural Networks for Feature Extraction}
\begin{figure}[!h]
	\includegraphics[width=\linewidth]{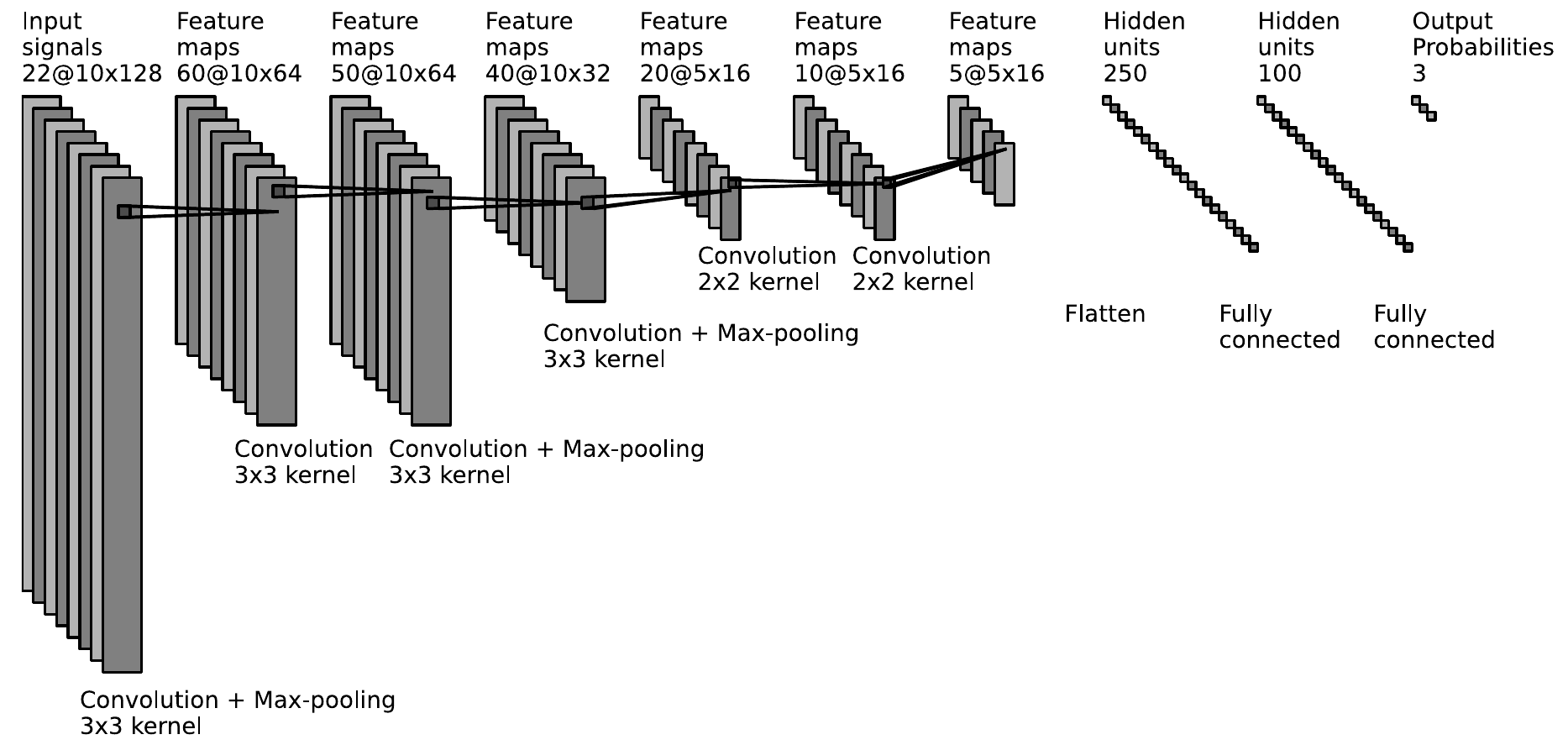}
	\caption{{\bf CNN architecture}
		Convolutional layers learn a set of filters (kernels) that are convolved with the output from the previous layer (feature map). Max-pooling layers downsample one or two dimensions of the feature map. Ex) The third convolutional layer learns a set of 50 kernels of size 3x3 and is followed by a pooling layer along the time dimension. Fully connected layers have a connection to every element of the output from the previous layer. Ex) the first fully connected layer has 250 units, each of these units has a weighted connection to the 5*5*16 elements in the previous feature map. Each unit in the output layer outputs a probability of belonging to one of the three classes. Generated with code from \url{http://www.github.com/gwding/draw_convnet}}
	\label{convnet}
\end{figure}

\paragraph*{\eatpunct}
CNN have demonstrated considerable success due to their ability to model local dependencies in the input and reduce the number of trained parameters in a neural network through weight sharing. We leverage both of these properties on the wavelet-transformed EEG. We use convolutional filters to learn features that capture the short term temporal dependencies of the EEG, and at the same time look for relationships between close frequency bands. We will briefly describe some features of deep CNN that we take advantage of here, for a full discussion see~\cite{Simard2003}.
\paragraph*{\eatpunct}
In addition to learning filter maps, CNN also feature two other common components; max-pooling and dropout. The max-pooling method allows the network to learn features that are temporally (or spatially) invariant. The network can then identify patterns in the coefficients of the wavelet tensor without considering whether the pattern occurs in the beginning or end of the signal slice. Dropout~\cite{Srivastava2014} randomly sets the output of units in the network to zero during training, preventing those units from affecting the output or the gradient of the loss function for an update step. This serves as a regularization technique to improve generalization by ensuring the network does not overfit by depending on specific hidden units.	
\paragraph*{\eatpunct}
The CNN trained had six convolutional layers followed by two dense layers, as shown in Fig~\ref{convnet}. The output layer consisted of three units with a soft-max activation representing a probability distribution over the three classes. We arrived at this architecture after experimenting with many different architectures, both shallower and wider. Deeper models proved difficult to train on the available hardware, while shallower models suffered from poor accuracy. The convolutional layers are stacked one after the other, decreasing the number of filters in each layer. Filter sizes were fixed at $3x3$ and $2x2$ as experiments with larger filters showed no improvement. Max-pooling layers were inserted after every other convolutional layer and dropout added after every layer. Instead of the typical sigmoid nonlinear activation, the rectified-linear unit is used for its non-saturating properties which inhibit the vanishing gradient problem~\cite{Nair2010}.	

\subsubsection{Aim 2: Computing the Length of the Preictal Period}
\paragraph*{\eatpunct}
The most crucial missing information in the seizure prediction problem is the actual length of the preictal period. Knowing the exact start and end points of the preictal period during the training process would greatly improve supervised machine learning approaches.	

\paragraph*{\eatpunct}
Since supervised machine learning algorithms require training data in the form of labeled examples of the incident of interest, missing or incorrect labeling poses a crippling problem. This is well studied in the machine learning community~\cite{Brodley1999}. In this context, the problem presents itself in two forms. (a) Assume the preictal period extends 8 minutes prior to seizure onset, but the labeler labels the period 15 minutes before seizure onset as belonging to the preictal class. Training a machine learning algorithm on such a dataset would result in very poor performance on the preictal class because the learner is given 7 minutes of signal that is interictal but labeled as preictal. This makes learning to differentiate between the interictal and preictal classes impossible. (b) If the opposite error is made and only 8 minutes are labeled as preictal but the preictal period extends 15 minutes before seizure onset the learner once again will fail. This is because the interictal class is now noisy with mislabeled examples from the preictal class. This is especially a problem in the context of automatic feature learning. Because the algorithm will find features that will perform well at the given task, in the presence of mislabeled data, the task the machine learns to do and the one we want it to do will not necessarily be the same.
\paragraph*{\eatpunct}
The literature provides some estimates to the length of the preictal period, but the estimates are not shown to be general~\cite{Osorio2011}. Our approach to deal with this problem is simple, we make the preictal length a parameter in the learning process and optimize over it using grid search.
To further validate the automatic selection of the preictal length, we analyze the extracted features from different preictal lengths.

\subsection{Methodology: A Prediction System}

\paragraph*{\eatpunct}
The input to the neural network is constructed by computing the wavelet transform on each recording to generate the tensors shown in Fig~\ref{wavelet_tensor}.  Following~\cite{Acar2007}, we use a set of didactic scales from 1 to 512 and the Mexican-hat mother wavelet. We introduce two parameters, epoch length and overlap percentage, and divide the tensor into overlapping windows of length $e$ seconds and overlapping by $o$ percent. Each of these windows becomes a separate training example to the neural network. We normalize by subtracting the mean and dividing by the standard deviation over each channel for each of the training examples.
\paragraph*{\eatpunct} 
In order to label the examples into the three classes; interictal, preictal, and ictal, we introduce an additional parameter $ l $ representing the assumed preictal length. Using the provided seizure onset time, we label all windows that fall $ l $ minutes before the seizure onset time as preictal and all windows after the seizure onset time as ictal (until the end of the recording). The remaining windows are considered interictal. 
\paragraph*{\eatpunct}
One of the common problems for machine learning algorithms is that they require balanced datasets in order to learn a non-trivial pattern from the data. The dataset is heavily imbalanced due to the ease of obtaining interictal data; the number of interictal examples outnumbers the other two classes by almost a factor of 8. Without balancing the classes, the classifier would achieve a low error rate by simply classifying all examples as interictal. This is a well-studied problem with many recommended techniques to balancing the dataset~\cite{Kubat1997}. Examples of these approaches are oversampling the minority classes, undersampling the majority classes, or combinations of both to balance the dataset~\cite{Chawla2002}. We adopt a simple undersampling scheme on the interictal class, randomly selecting a number of majority class examples such that the classes are balanced.
\paragraph*{\eatpunct}
Before training the CNN on the labeled data, we split the data into a training and validation set using k-fold crossvalidation. We set k to 10 and split the data into 90\% for training data, and 10\% for validation data. We use the validation set for hyperparameter optimization and monitor the loss function on the validation set as the criteria for early stopping. Some of the important hyperparameters to optimize include; epoch length, overlap percentage, and preictal length. Early stopping is used to prevent overfitting the training dataset by halting training when loss on the validation set begins to increase.
\paragraph*{\eatpunct}
Training the deep convolutional network was done using stochastic gradient descent with an adaptive learning rate~\cite{Zeiler2012}. Gradients were estimated using the Keras~\cite{chollet2015} wrapper for the Theano library~\cite{TheanoDevelopmentTeam2016}. It is important to note that the loss function for training was not based on seizure prediction performance (the network's ability to detect the seizure before it occurred). Instead the network was trained using the categorical cross-entropy loss function over the three classes.
\paragraph*{\eatpunct}
After training is complete, the network enters the inference stage where it generates a probability distribution over the three classes for a new input signal. We are interested in when the patient leaves the interictal state and use the output probabilities of the network to model this. Given $ p_0 $, $ p_1 $, and $ p_2 $ as the output probabilities for the interictal, preictal, and ictal classes respectively, where $ p_0+p_1+p_2=1 $, we compute the probability of an oncoming seizure as $ p=p_1+p_2=1-p_0 $.
\paragraph*{\eatpunct} 
Until this point, each input window has been treated independently. We introduce dependence between adjacent windows by allowing a previous prediction to influence the current one and generate smooth outputs:
$$s(0)=p(0)$$
$$s(t)=\alpha p(t)+(1-\alpha) s(t-1)$$
Where $\alpha$ is the exponential smoothing parameter. This effectively smooths the output over time and allows us to control the sensitivity of the predictor. The system declares a seizure imminent (within minutes) when the signal $ s $ crosses a threshold. We show results using an empirically determined threshold of $0.6$ but evaluate the performance of the system using the area under the ROC curve (ROC-AUC).
\paragraph*{\eatpunct}
The output of the last hidden layer of the trained network, the 100-unit fully connected layer of the network shown in Fig~\ref{convnet} represents a compressed version of the input signal. We can observe the output of each unit in this layer for each epoch, yielding a $T$ x $100$ matrix of extracted features. This is useful for analyzing the phase transition between the interictal and preictal state.
\paragraph*{\eatpunct}
Hyperparameter optimization was done using a grid search on a range of values for each variable. The final set of hyperparameters was chosen as the set achieving the lowest average validation loss over the 10 fold cross-validation. The final hyperparameters after training are a $e=1$ second epoch length, $ o=0\% $ overlap between windows, $ \alpha=0.7 $, and $ l=10 $ minute preictal length. The preictal length optimized through cross-validation in the training phase is set as the prediction horizon for the system.

\section{Results}
\paragraph*{\eatpunct}
In addition to describing the performance of the system on the test datasets in this section, we verify that the features learned by the CNN capture the interictal to preictal phase transition using the Kullback-Leibler (KL) divergence.

\subsection{Extracted Feature Analysis for Phase Transition}
\begin{figure}[!h]
	\includegraphics[width=\linewidth]{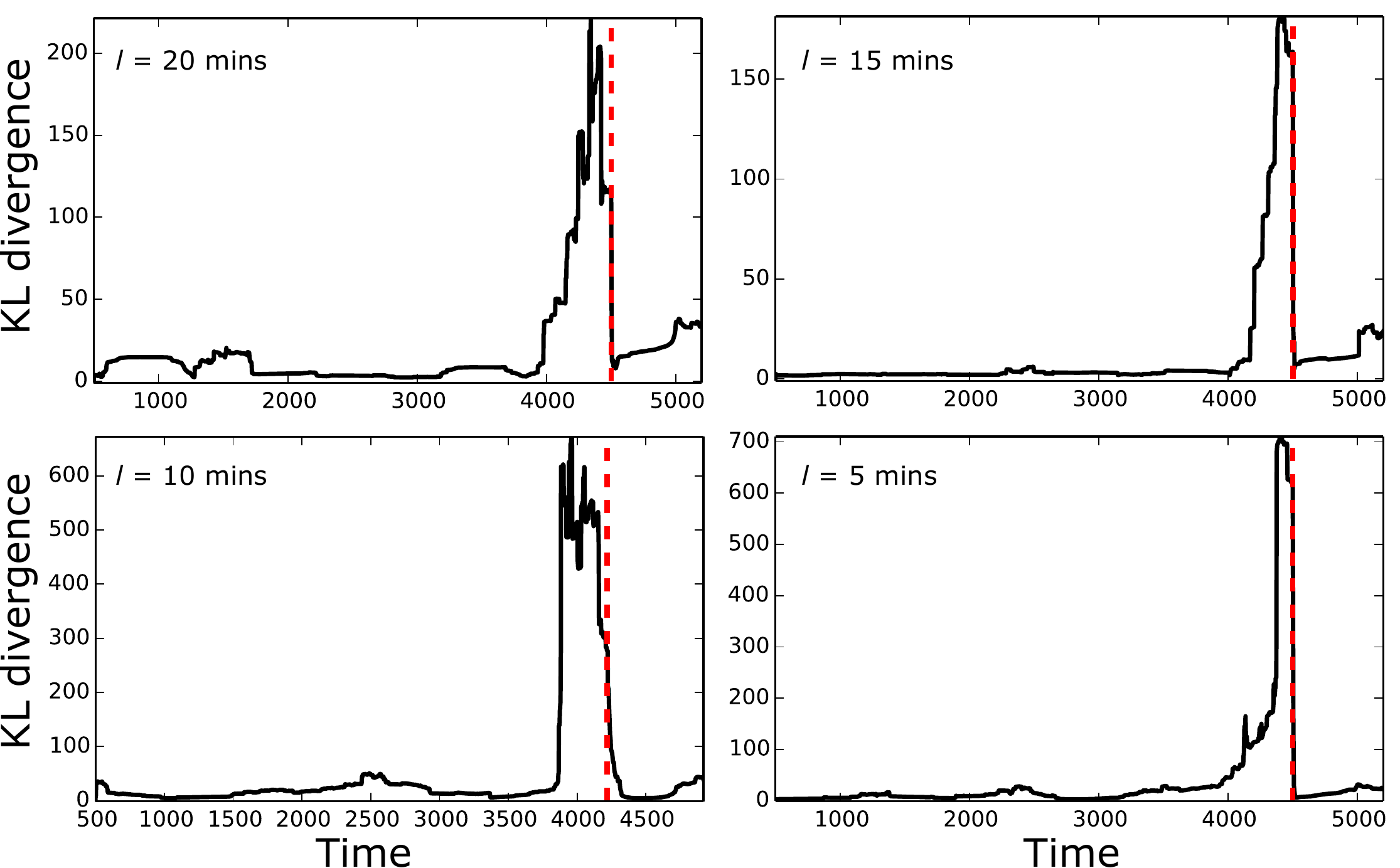}
	\caption{{\bf Verifying the preictal period length}
		KL-divergence between interictal feature distribution and feature distributions over a single recording for preictal lengths of 20, 15, 10, and 5 minutes. Notice the abrupt increase in divergence at around 9-10 mins before seizure onset (onset marked with red dashed line) irrespective of the preictal length parameter value. This pattern was observed across recordings indicating a approximate preictal length of 10 minutes.}
	\label{kldiv}
\end{figure}

\paragraph*{\eatpunct}
The features extracted by the network under different preictal lengths are analyzed to verify the preictal length value yielded by cross-validation. Specifically, we want to find the earliest time before seizure onset where a dramatic change in the distribution of extracted features occurs (i.e. a change point). After selecting a subset of the features and de-correlating them, two distributions are estimated in the extracted features; the distribution during interictal-only periods and the distribution around a time point $ t $. The interictal distribution is approximated by a multivariate Gaussian with mean $ \mu_0 $ and covariance $ \Sigma_0 $ calculated from the features of epochs in the interictal period. Similarly, we approximate the distribution of features at $ t $ by calculating the mean $ \mu_1 $ and covariance $ \Sigma_1 $ of the set of features from $ t-L $ to $ t $, where $ L $ is the number of samples. We then measure the divergence between the interictal distribution and the distribution at $ t $ using the following form of the KL divergence with $ k $ set as the number of features:
$$ D(t)=\frac{1}{2} (tr(\Sigma_1^{-1} \Sigma_0) + {(\mu_1-\mu_0) }^T \Sigma_1^{-1} (\mu_1-\mu_0)-k+ \ln{\frac{det{\Sigma_1}}{det{\Sigma_0}}}) $$
Using this measure with $\mu_0, \mu_1, \Sigma_0,$ and $\Sigma_1$ from above, the earliest time is found where the baseline and current distributions are significantly different; Fig~\ref{kldiv} shows an example of this measure computed on a recording from the MSSM dataset for four preictal lengths. The divergence is small and constant during the first part of the recording but increases rapidly at approximately 9-10 minutes before the seizure onset. This means that there is a shift in the distribution of extracted features at that time. Furthermore, the shift occurs at the same location irrespective of the value of the preictal length parameter, indicating the features which are learned are capturing the transition in the underlying system. The same pattern also appears consistently in other recordings from the validation set. This affirms the prediction horizon determined using cross-validation and indicates the phase transition from the interictal state to the preictal state that is captured by these features occurs around 10 minutes before seizure onset.

\subsection{Test Set Results}
\begin{figure}[!h]
	\includegraphics[width=\linewidth]{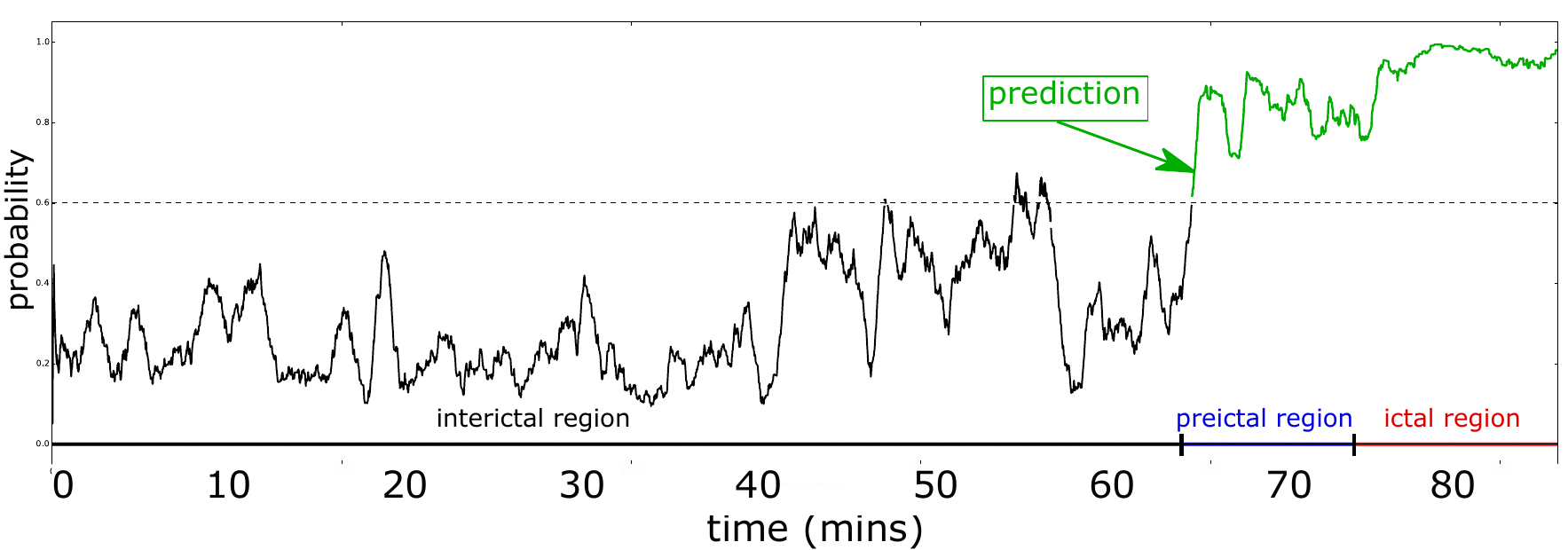}
	\caption{{\bf Seizure recording example}
		Oncoming seizure probability shown for a seizure (ictal) recording. The colored horizontal bar directly above the x-axis labels the interictal (black), preictal (blue), and ictal (red) regions of the signal. The curve is the probability output generated by the CNN, where the green coloring indicates the system predicts an oncoming seizure; determined by a significant crossing of the threshold (dotted line).}
	\label{ictal_raw}
\end{figure}

\paragraph*{\eatpunct}
The results on the MSSM and CHB-MIT test sets are summarized in Table~\ref{table2} and Table~\ref{table3} respectively. Table~\ref{table6} lists the number of seizures and the total length of the recordings for each patient in the test set. Table~\ref{table5} compares the specificity and sensitivity results for this work and three other seizure prediction methods. Sensitivity is defined as the percentage of seizures correctly predicted. Specificity is given by the false prediction rate (FPr) which is the number of false predictions divided by the total length of interictal-only periods. The prediction times of seizures from each patient tested aligned by the seizure onset time are shown in Fig~\ref{horizon}. 
Fig~\ref{ictal_raw}
 shows an example of the predictions generated by the system on a seizure recording using a fixed threshold ($\tau = 0.6$).
\paragraph*{\eatpunct}
In addition to measuring seizure prediction performance, we trained an identical CNN on the raw data by feeding the raw EEG signal to the network (omitting the wavelet transform step) to evaluate the advantages of training on wavelet-transformed data. Comparison of the two networks on the test set using Matthews correlation coefficient~\cite{Matthews1985a} demonstrated that using wavelet-transformed EEG as input to the network resulted in improved performance across all test set recordings. One reason for this improvement is the filtering effect of the wavelet transform allows the CNN to learn features that involve activity across multiple frequency bands.

\subsubsection{MSSM Test Set}
\paragraph*{\eatpunct}

Three recordings in this test set are interictal-only from three different patients. The remaining 15 recordings contain seizures from 12 different patients. The average prediction time for the seizures was 8 minutes before seizure onset. One of the 15 seizures tested was not predicted by the system. The false prediction rate was .128/hr.  Interestingly, false positives did not occur in interictal-only recordings but occurred in seizure recordings long before the 10 minute prediction horizon.

% Table 2
\begin{table}[!ht]

	\centering
	\caption{
		{\bf Results reported on the 18 MSSM test set recordings. Prediction time before seizure onset (if any) are shown, in addition to any false predictions raised by the system. Measurements are given with respect to seizure onset time. A false prediction is recorded when the system reports a seizure oncoming outside of the prediction horizon (10 minutes).}}
	\begin{tabular}{|l|l+l|l|l|l|}
		\hline
		{\bf } &  & {\bf Pred.} & {\bf False}  &  \\ 
		{\bf Patient \#} & {\bf Type}  & {\bf time (secs)} & {\bf pred.} & {\bf ROC-AUC} \\ \thickhline
		2 & interictal & N/A & 0 & N/A  \\ \hline
		4 & interictal & N/A & 0 & N/A  \\ 				 
		 & left temporal & 460 & 0 & 0.935 \\ \hline
		5 & right temporal & 557 & 0 & 0.961 \\ \hline
		6 & interictal & N/A & 0 & N/A  \\
		 & right temporal & 452 & 0 & 0.852  \\	\hline
		11 & left temporal & 586 & 0 & 0.941 \\ 
		 & left temporal & 410 & 0 & 0.903 \\ \hline		
		 & left frontotemporal & -35 & 0 & 0.448 \\ \hline
		18 & left temporal & 234 & 0 & 0.795 \\ 
		 & left temporal & 515 & 0 & 0.955 \\ 
		 & left temporal & 541 & 0 & 0.891 \\ \hline
		19 & left temporal & 512 & 0 & 0.934 \\ \hline
		22 & right frontotemporal & 577 & 1 & 0.949 \\ \hline
		24 & left temporal & 569 & 1 & 0.934 \\ \hline
		25 & right temporal & 532 & 0 & 0.945 \\ \hline
		26 & right temporal & 536 & 0 & 0.937 \\ \hline
		27 & left temporal & 520 & 1 & 0.903 \\ \hline
		Avg. &  		& 464.4 & & 0.885 \\ \hline
	\end{tabular}
	\label{table2}
\end{table}

\subsubsection{CHB-MIT Test Set}
\paragraph*{\eatpunct}
The CHB-MIT test set consists of 18 seizure recordings and 50 interictal recordings, allowing us to evaluate the specificity and false prediction rate of the system accurately. The total of 68 recordings corresponds to approximately 70.5 hours of EEG, of which 50 hours are interictal only. 
%The results are summarized in Table~\ref{table3}, 
Seizures were predicted on average 6 minutes before the seizure onset time with three of the 18 seizures not predicted. The false prediction rate was .147/h. In this set of recordings, false predictions occurred within interictal only recordings.

% Table 3
\begin{table}[!ht]
	\centering
	\caption{
		{\bf Results reported on the 18 CHB-MIT test set seizure recordings (50 interictal recordings not shown). Prediction time before seizure onset (if any) are shown, in addition to any false predictions raised by the system. Measurements are given with respect to seizure onset time. A false prediction is recorded when the system reports a seizure oncoming outside of the prediction horizon (10 minutes). (* More information about the dataset can be found at \url{https://www.physionet.org/pn6/chbmit/})}}
	\begin{tabular}{|l|l+l|l|l|l|}
		\hline
		 &  & {\bf Pred.} & {\bf False}  &  \\ 
		{\bf Patient } & {\bf Type*}  & {\bf time (secs)} & {\bf pred.} & {\bf ROC-AUC} \\ \thickhline
		chb01 & seizure & 546 & 0 & 0.943  \\ \hline
		chb02 & seizure & 372 & 0 & 0.855  \\ \hline
		chb03 & seizure & 392 & 0 & 0.853  \\ \hline
		chb04 & seizure & 554 & 0 & 0.973  \\ \hline		 
		chb05 & seizure & 317 & 0 & 0.823 \\
			  & seizure & 551 & 0 & 0.988 \\ \hline
		chb07 & seizure & 385 & 0 & 0.858 \\ \hline
		chb08 & seizure & 470 & 0 & 0.943 \\ 
			  & seizure & 488 & 0 & 0.921 \\ \hline
		chb10 & seizure & 313 & 0 & 0.855 \\ \hline
		chb11 & seizure & 445 & 0 & 0.878 \\ \hline
		chb13 & seizure & 224 & 0 & 0.802 \\\hline
		chb17a & seizure & 532 & 0 & 0.966 \\ 
			   & seizure & 475 & 0 & 0.940 \\ \hline
		chb19 & seizure & -36 & 0 & 0.713 \\ \hline
		chb21 & seizure & -73 & 0 & 0.699 \\ \hline
		chb22 & seizure & 389 & 0 & 0.877 \\ \hline
		chb24 & seizure & -46 & 0 & 0.708 \\ \hline
		Avg. &  		& 349.9 & & 0.866 \\ \hline
	\end{tabular}
	\label{table3}
\end{table}

\begin{figure}[!h]
	\includegraphics[width=\linewidth]{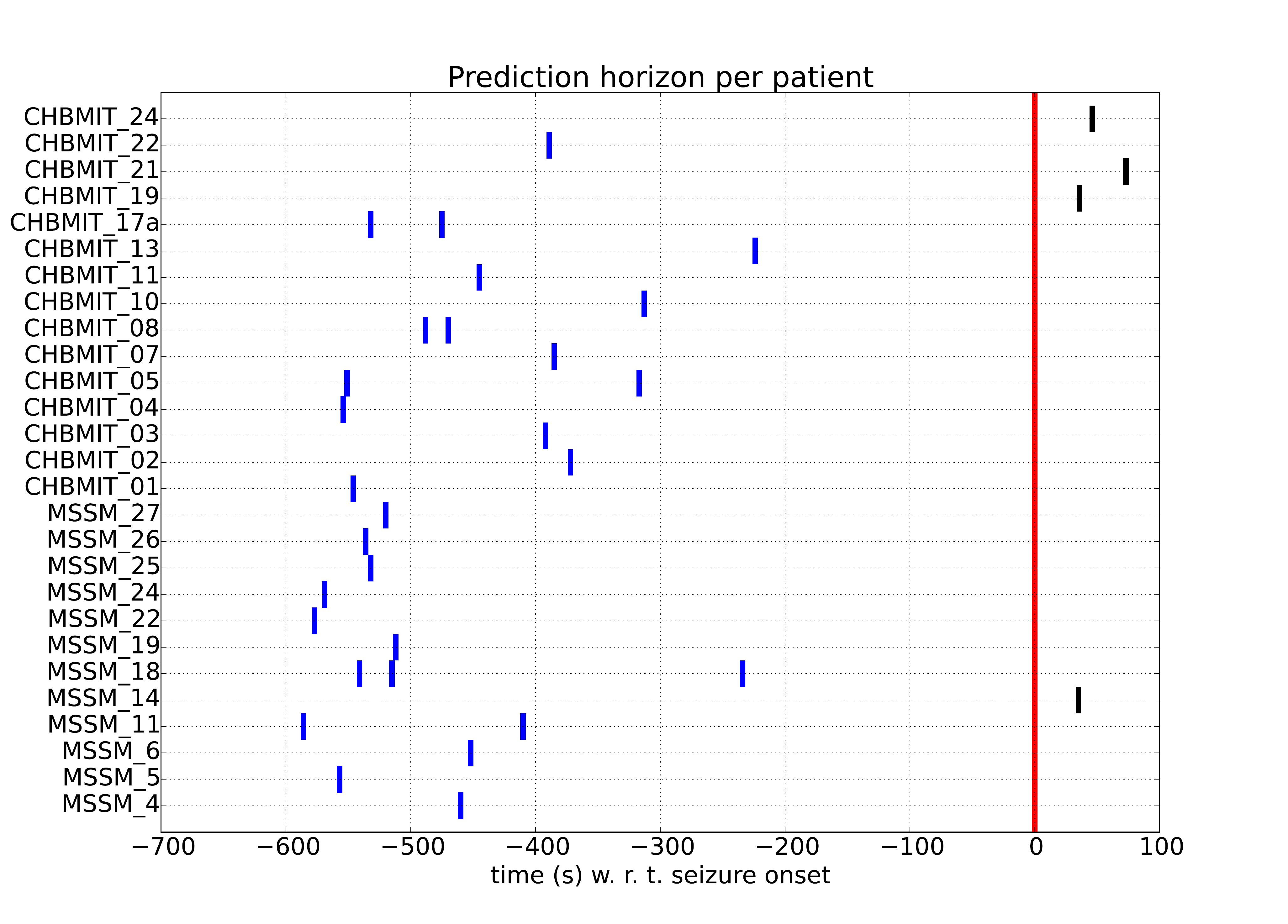}
	\caption{{\bf Prediction horizons for test patients}
		The prediction time of the system is shown for all test set recordings. Each row shows the prediction time the system achieved for all the recordings from a single patient. The vertical red line indicates the seizure onset time. Predictions colored black were reported after the seizure occurred. The standard deviation of prediction times across all the recordings is large ($\sigma=98.7$). This indicates that while the preictal phase transition is captured by the system for most patients, a patient-specific system could decrease the spread of prediction times.}
	\label{horizon}
\end{figure}

\begin{table}[!ht]
	\centering
	\caption{
		{\bf Number of test seizures and recording lengths for each patient}}
	\begin{tabular}{|l|l+l|l|l|l|}
		\hline
		{\bf Patient} & {\bf \# of seizures} & {\bf Length (mins)}  \\
		 \thickhline
 		6 & 1 & 85 \\			 
 		4 & 1 & 85 \\ 
 		11 & 2 & 170  \\ 
 		5 & 1 & 85  \\ 
 		14 & 1 & 85  \\ 
 		18 & 3 & 255  \\ 
 		19 & 1 & 85  \\ 
 		22 & 1 & 85  \\ 
 		24 & 1 & 85  \\ 
 		25 & 1 & 85  \\ 
 		26 & 1 & 85  \\ 
 		27 & 1 & 85  \\
		chb01 & 1 & 60  \\
		chb02+ & 1 & 60  \\ 
		chb03 & 1 & 60   \\ 
		chb04 & 1 & 60 \\			 
		chb05 & 2 & 120 \\ 
		chb07 & 1 & 60 \\ 
		chb08 & 2 & 120 \\ 
		chb10 & 1 & 60 \\ 
		chb11 & 1 & 60 \\ 
		chb13 & 1 & 60 \\
		chb17a & 2 & 120 \\ 
		chb19 & 1 & 60  \\ 
		chb21 & 1 & 60  \\ 
		chb22 & 1 & 60  \\ 
		chb24 & 1 & 60 \\ \hline
	\end{tabular}
	\label{table6}

\end{table}

\subsection{Error Analysis}
\begin{figure}[!h]
	\includegraphics[width=\linewidth]{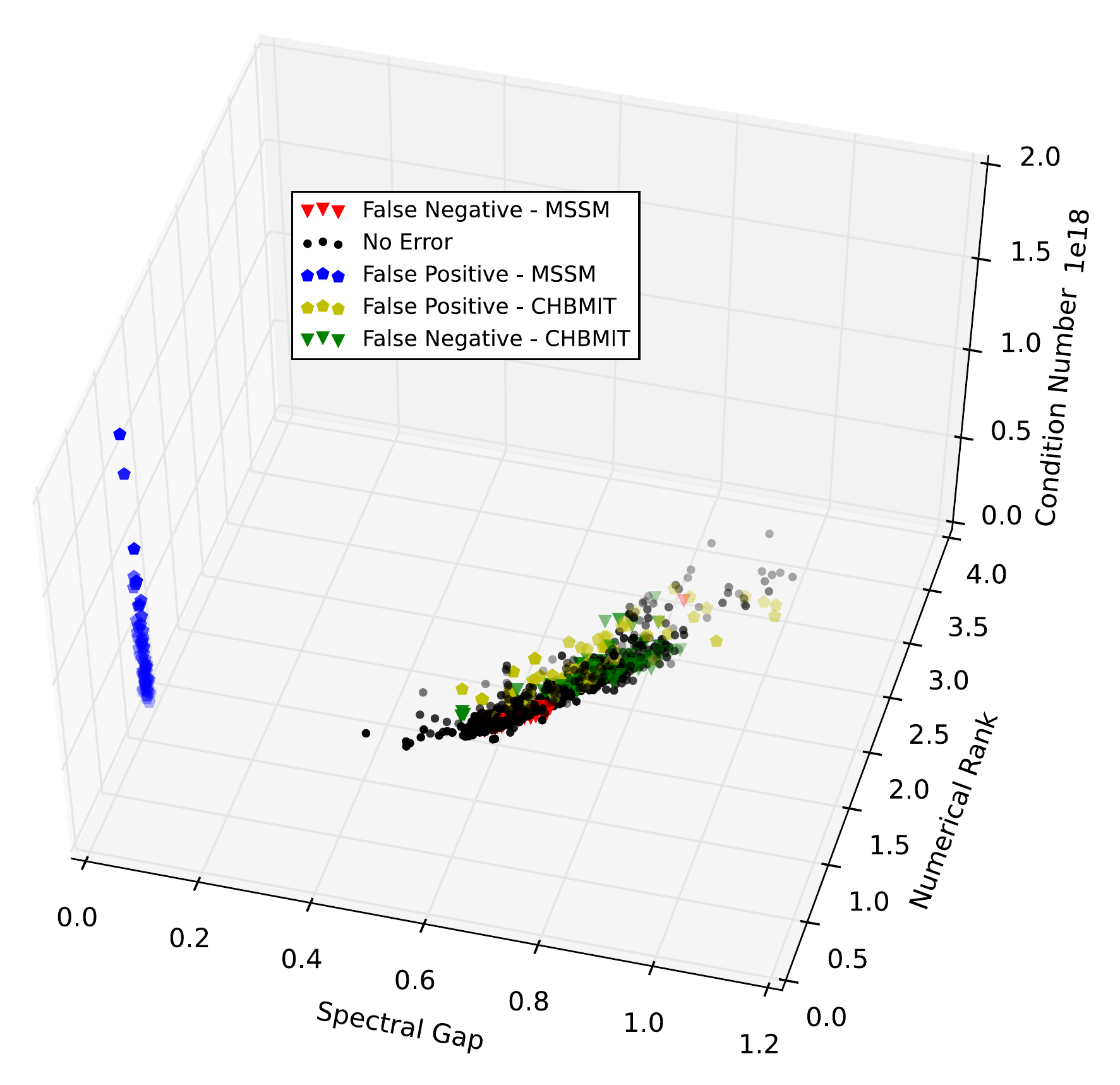}
	\caption{{\bf Error analysis}
		Scatterplot showing spectral gap, numerical rank, and condition number of each channel of each test set recording. Points are colored based on the dataset and the type of error observed. We observed that some recordings from the MSSM test set are poorly conditioned and resulted in false positives. However, other recordings that resulted in false positives/negatives are indistinguishable in this space from recordings without errors.}
	\label{error_analysis}
\end{figure}

\paragraph*{\eatpunct}
Analysis of the raw EEG signal revealed some causes of error, particularly false positives. This can be seen in the linear algebraic properties of the wavelet matrices obtained by slicing the wavelet tensor shown in Fig~\ref{wavelet_tensor} along the “channels” mode. Each slice is a $T$ x $10$ matrix of wavelet coefficients. In Fig~\ref{error_analysis}, we show a scatterplot of the spectral gap, numerical rank, and condition number of these matrices. The spectral gap of a matrix can be calculated using the non-zero singular values $\sigma_1$, $\sigma_2$, ..., $\sigma_r$ of the matrix, with $r \leq 10$. Spectral gap is given by the ratio of the first and second singular values:
$$ \frac{\sigma_2}{\sigma_1} $$
The numerical rank of the matrices is estimated by computing the squared ratio of the Frobenius norm and the spectral norm of the matrix $A$:
$$ \frac{||A||_F^2}{||A||_2^2} $$
The condition number of a matrix is given by the ratio of the largest singular value and the smallest singular value:
$$ \frac{\sigma_1}{\sigma_r} $$
\paragraph*{\eatpunct}
We observed that the false positives incurred on the MSSM test set are due to the poor conditioning of the matrices. This is indicated by the low spectral gap and high condition numbers for those recordings, shown as blue pentagons in Fig~\ref{error_analysis}. False positives from the CHBMIT dataset did not follow this pattern and are indistinguishable from recordings without error in this space. False negatives from both datasets are also scattered among recordings without error indicating the underlying causes are not captured by a linear analysis.
\paragraph*{\eatpunct}
The four seizures not predicted by the system were also visually inspected.  From the MSSM database, the single seizure not predicted was in a low amplitude EEG ($ \leq 15$ uV) with some muscle artifact superimposed.  From the CHB-MIT database, two of the seizures had substantial electrode and movement artifact, likely making prediction impossible.  The third seizure not predicted had numerous sleep/wake transitions and frequent epileptiform bursts. Sleep wake transitions and epileptiform bursts were also found in numerous EEGs in which the program was successful.  Further data of failed studies may be helpful in refining the program. 

\subsection{Sensitivity and Comparison to Unspecific Random Predictor}
\paragraph*{\eatpunct}
The sensitivity of the system is shown in Table~\ref{table5}, alongside the sensitivity range of an unspecific random predictor ($\sigma_{low} - \sigma_{up}$) ~\cite{Schelter2006}. The unspecific random predictor is used as a baseline to statistically validate the sensitivity reported by a model as significant. The sensitivity of the analytic random predictor is based on the performance of a predictor that uses a Poisson process to generate predictions. This is determined by a number of parameters; the seizure occurrence period (SOP - defined as the duration after a prediction in which a seizure must occur in order for the prediction to be correct), a fixed false prediction rate (FPr), and the number of independent features analyzed ($d$). The sensitivity range of the random predictor is given by (where $K$ is the number of analyzed seizures)~\cite{Schelter2006}:
$$ P = \text{SOP} * \text{FPr} $$
$$ \sigma_{low} = \max_k\left(1 - \left( \sum_{j<k}{{K \choose j}P^j(1-P)^{K-j}}\right)\right) $$
$$ \sigma_{up} = \max_k\left(1 - \left( \sum_{j<k}{{K \choose j}P^j(1-P)^{K-j}}\right)^d\right)$$
The ranges in the last column of Table~\ref{table5} were calculated using SOP $=10$ minutes, FPr set to the observed false prediction rates in Table~\ref{table5}, and a significance level $\alpha=0.05$. The number of independent features $d$ is difficult to estimate and was set to the upper bound of $d=100$ (corresponding to the number of units in the second to last layer of the CNN).
% Table 4
\begin{table}[!ht]
	\centering
	\caption{
		{\bf Specificity and false prediction rate (FPr) for this work, the algorithm proposed by Cook et al. described in the appendix of~\cite{Cook2013b}, and two of the top three seizure prediction algorithms from Kaggle~\cite{Brinkmann2016}. Comparison to an unspecific random predictor~\cite{Schelter2006} with $\sigma_{low}$ and $\sigma_{high}$ calculated using SOP=10 mins, recorded FPr, and $\alpha=0.05$. A sensitivity above the $\sigma_{high}$ value indicates the algorithm performs better than chance at the given FPr. Kaggle1 algorithm code: \url{https://github.com/MichaelHills/seizure-prediction}. Kaggle2 algorithm code: \url{https://github.com/jlnh/SeizurePrediction}.}}
	\begin{tabular}{|l+l|l|l|l|}
		\hline
		{\bf Method} & {\bf Sensitivity}  &  {\bf FPr} & {\bf Random pred.} \\ 
					 & 				& {\bf (FP/h)} & {\bf $\sigma_{low}$} - {\bf $\sigma_{high}$} \\ \hline
		Kaggle1~\cite{Brinkmann2016} & 72.7\% & 0.285 & 15.1\% - 27.2\% \\ 
		Kaggle2~\cite{Brinkmann2016} & 75.8\% & 0.230 & 12.1\% - 24.2\% \\
		Cook et al.~\cite{Cook2013b} & 66.7\% & 0.186 & 12.1\% - 21.2\% \\
		{\bf This work}  & {\bf 87.8\%} & {\bf 0.142} &  9.1\% - 15.1\% \\ \hline
	\end{tabular}
	\label{table5}

\end{table}

\section{Computational Comparison to Prior Work}
\paragraph*{\eatpunct}
 Testing the learned features on an out-of-sample dataset achieves results with sensitivity and specificity superior to other published seizure prediction methods~\cite{Mormann2007,Gadhoumi2016a}. We evaluated the performance of three other algorithms on our datasets (accessible online \url{http://www.dsrc.rpi.edu/?page=databank}). Two of the algorithms, Kaggle1 and Kaggle2 in Table~\ref{table5}, are top performers from the Kaggle seizure prediction competition~\cite{Brinkmann2016} (\url{https://www.kaggle.com/c/seizure-prediction}). Participants were asked to develop algorithms that could distinguish between interictal and preictal EEG samples. The Kaggle1 algorithm uses a combination of time and frequency domain features, such as the spectral entropy of selected frequency bands, time correlation matrix, frequency correlation matrix, and higuchi fractal dimension, with logistic regression as a predictor. The Kaggle2 algorithm uses a similar approach but with fewer features and a support vector machine classifier. The third algorithm evaluated is the one proposed by Cook et al.~\cite{Cook2013b}, which calculates average energy, Teager-Kaiser energy, and line-length on an array of six filters over each channel. After a feature selection step, a hybrid decision-tree/k-nearest neighbor classifier is used for classification. Since we are not aware of any published codes for the algorithm, we implemented the algorithm in Python. As shown in Table~\ref{table5}, each of the methods performed better than an unspecific random predictor. Our system achieves higher sensitivity (87.8\%) than all of the other methods with a lower false prediction rate (0.142 FP/h). 
\section{Discussion}
\paragraph*{\eatpunct}
Our results show that a preictal phase transition can be observed in scalp EEG data using features automatically extracted from the signal. The results also showed that typical EEG artifacts and changes, such as sleep-wake transitions, did not cause false positives. In addition, the phase transition is shown to occur approximately 10 minutes before seizure onset as reported in previous studies~\cite{Aarabi2012}. However, as shown in Fig~\ref{horizon}, the spread of prediction time of the preictal transition is large ($\sigma=98.7$). This spread is a drawback of training on data from many different patients, which was necessary because of the limited amount of patient specific data.
\paragraph*{\eatpunct}
The datasets used in this study contain only scalp EEG recordings which, as mentioned previously, are easier to obtain but suffer drawbacks in signal quality. Applying our methodology to a dataset of intracranial EEG recordings could allow  the system to learn features that detect the preictal phase transition further in advance and with less variability, similar to studies using intracranial EEG which report a prediction horizon of an hour or longer~\cite{Mormann2007}.

\section{Conclusion}
\paragraph*{\eatpunct} 
These results suggest real feasibility in creating a reliable seizure prediction system.  The goal would be to create a wearable non-invasive EEG device which would alert patients, family members, and doctors to imminent seizures. This has the potential to enhance the safety of patients, decrease the rates of sudden unexpected death in epilepsy patients (SUDEP) and perhaps to allow some patients to take medications only when needed and not chronically.
\paragraph*{\eatpunct}
To achieve a complete system, a number of hurdles still need to be overcome. The limited database of scalp EEG recordings needs to be extended in order to fully test the generalization ability of the system. In addition, a clear understanding of how the ictal state ends is required. This information will allow the predictor to “reset” itself after an ictal event. Creating an end to end deep neural network that directly optimizes over the prediction task is another step that we intend to pursue. 

\section*{Acknowledgments}
We thank Nimit Dhulekar and the reviewers for valuable discussions and comments. None of the authors have potential conflicts of interest to be disclosed. This work was supported in part by NSF Award \#1302231.

\ifCLASSOPTIONcaptionsoff
  \newpage
\fi


\Urlmuskip=0mu plus 1mu\relax
\bibliographystyle{myIEEEtran}
\bibliography{IEEEabrv,library}

\begin{thebibliography}{10}
\providecommand{\url}[1]{#1}
\csname url@samestyle\endcsname
\providecommand{\newblock}{\relax}
\providecommand{\bibinfo}[2]{#2}
\providecommand{\BIBentrySTDinterwordspacing}{\spaceskip=0pt\relax}
\providecommand{\BIBentryALTinterwordstretchfactor}{4}
\providecommand{\BIBentryALTinterwordspacing}{\spaceskip=\fontdimen2\font plus
\BIBentryALTinterwordstretchfactor\fontdimen3\font minus
  \fontdimen4\font\relax}
\providecommand{\BIBforeignlanguage}[2]{{%
\expandafter\ifx\csname l@#1\endcsname\relax
\typeout{** WARNING: IEEEtran.bst: No hyphenation pattern has been}%
\typeout{** loaded for the language `#1'. Using the pattern for}%
\typeout{** the default language instead.}%
\else
\language=\csname l@#1\endcsname
\fi
#2}}
\providecommand{\BIBdecl}{\relax}
\BIBdecl

\bibitem{Shrikumar2016}
A.~Shrikumar \emph{et~al.}, ``{Not Just a Black Box: Interpretable Deep
  Learning by Propagating Activation Differences},'' \emph{arXiv}, vol.~1,
  no.~3, pp. 0--5, 2016.

\bibitem{Gadhoumi2016a}
K.~Gadhoumi \emph{et~al.}, ``{Seizure prediction for therapeutic devices: A
  review},'' \emph{Journal of Neuroscience Methods}, vol. 260, no. 029, pp.
  270--282, 2016.

\bibitem{Harrison2005}
M.~A.~F. Harrison \emph{et~al.}, ``{Correlation dimension and integral do not
  predict epileptic seizures},'' \emph{Chaos}, vol.~15, no.~3, p. 033106, sep
  2005.

\bibitem{Freestone2015}
D.~R. Freestone \emph{et~al.}, ``{Seizure Prediction: Science Fiction or Soon
  to Become Reality?}'' \emph{Current Neurology and Neuroscience Reports},
  vol.~15, no.~11, p.~73, nov 2015.

\bibitem{Brinkmann2016}
B.~H. Brinkmann \emph{et~al.}, ``{Crowdsourcing reproducible seizure
  forecasting in human and canine epilepsy},'' \emph{Brain}, vol. 139, no.~6,
  pp. 1713--1722, jun 2016.

\bibitem{Cook2013b}
M.~J. Cook \emph{et~al.}, ``{Prediction of seizure likelihood with a long-term,
  implanted seizure advisory system in patients with drug-resistant epilepsy: A
  first-in-man study},'' \emph{The Lancet Neurology}, vol.~12, no.~6, pp.
  563--571, 2013.

\bibitem{freiburg2010data}
E.~E.~G. Freiburg, ``{Data Base},'' \emph{Epilepsy center of the university
  hospital of Freiburg}, 2010.

\bibitem{Ihle2012}
M.~Ihle \emph{et~al.}, ``{EPILEPSIAE - A European epilepsy database},''
  \emph{Computer Methods and Programs in Biomedicine}, vol. 106, no.~3, pp.
  127--138, 2012.

\bibitem{Cherkassky2015}
V.~Cherkassky \emph{et~al.}, ``{Reliable seizure prediction from EEG data},''
  in \emph{2015 International Joint Conference on Neural Networks (IJCNN)},
  vol. 2015-Septe.\hskip 1em plus 0.5em minus 0.4em\relax IEEE, jul 2015, pp.
  1--8.

\bibitem{Korshunova2017}
I.~Korshunova \emph{et~al.}, ``{Towards improved design and evaluation of
  epileptic seizure predictors},'' \emph{IEEE Transactions on Biomedical
  Engineering}, vol. 9294, no.~c, pp. 1--1, 2017.

\bibitem{Chisci2010}
L.~Chisci \emph{et~al.}, ``{Real-time epileptic seizure prediction using AR
  models and support vector machines},'' \emph{IEEE Transactions on Biomedical
  Engineering}, vol.~57, no.~5, pp. 1124--1132, 2010.

\bibitem{Karoly2016}
P.~J. Karoly \emph{et~al.}, ``{Interictal spikes and epileptic seizures: Their
  relationship and underlying rhythmicity},'' \emph{Brain}, vol. 139, no.~4,
  pp. 1066--1078, 2016.

\bibitem{Li2013}
S.~Li \emph{et~al.}, ``{Seizure prediction using spike rate of intracranial
  EEG},'' \emph{IEEE Transactions on Neural Systems and Rehabilitation
  Engineering}, vol.~21, no.~6, pp. 880--886, 2013.

\bibitem{Mormann2007}
F.~Mormann \emph{et~al.}, ``{Seizure prediction: The long and winding road},''
  \emph{Brain}, vol. 130, no.~2, pp. 314--333, 2007.

\bibitem{Nagaraj2015}
V.~Nagaraj \emph{et~al.}, ``{Future of Seizure Prediction and Intervention :
  Closing the Loop},'' \emph{Journal of clinical neurophysiology : official
  publication of the American Electroencephalographic Society}, vol.~32, no.~3,
  pp. 194--206, jun 2015.

\bibitem{LeVanQuyen1999}
M.~{Le Van Quyen} \emph{et~al.}, ``{Anticipating epileptic seizure in real time
  by a nonlinear analysis of similarity between {\{}EEG{\}} recordings},''
  \emph{NeuroReport}, vol.~10, p. 2149, 1999.

\bibitem{Iasemidis2003}
L.~D. Iasemidis \emph{et~al.}, ``{Adaptive epileptic seizure prediction
  system.}'' \emph{IEEE transactions on bio-medical engineering}, vol.~50,
  no.~5, pp. 616--627, 2003.

\bibitem{Cho2017a}
D.~Cho \emph{et~al.}, ``{EEG-Based Prediction of Epileptic Seizures Using Phase
  Synchronization Elicited from Noise-Assisted Multivariate Empirical Mode
  Decomposition},'' \emph{IEEE Transactions on Neural Systems and
  Rehabilitation Engineering}, vol.~25, no.~8, pp. 1309--1318, 2017.

\bibitem{Chu2017}
H.~Chu \emph{et~al.}, ``{Predicting epileptic seizures from scalp EEG based on
  attractor state analysis},'' \emph{Computer Methods and Programs in
  Biomedicine}, vol. 143, pp. 75--87, 2017.

\bibitem{Aarabi2012}
A.~Aarabi and B.~He, ``{A rule-based seizure prediction method for focal
  neocortical epilepsy},'' \emph{Clinical Neurophysiology}, vol. 123, no.~6,
  pp. 1111--1122, jun 2012.

\bibitem{Duncan2013}
D.~Duncan and R.~Talmon, ``{Identifying preseizure state in intracranial EEG
  data using diffusion kernels},'' \emph{Mathematical Biosciences and
  engineering}, vol.~00, pp. 1--13, 2013.

\bibitem{Dhulekar2014}
N.~Dhulekar \emph{et~al.}, ``{Graph-theoretic analysis of epileptic seizures on
  scalp EEG recordings},'' \emph{Proceedings of the 5th ACM Conference on
  Bioinformatics, Computational Biology, and Health Informatics - BCB '14}, pp.
  155--163, 2014.

\bibitem{Schelter2006}
B.~Schelter \emph{et~al.}, ``{Testing statistical significance of multivariate
  time series analysis techniques for epileptic seizure prediction},''
  \emph{Chaos: An Interdisciplinary Journal of Nonlinear Science}, vol.~16,
  no.~1, p. 13108, 2006.

\bibitem{Aarabi2014}
A.~Aarabi and B.~He, ``{Seizure prediction in hippocampal and neocortical
  epilepsy using a model-based approach},'' \emph{Clinical Neurophysiology},
  vol. 125, no.~5, pp. 930--940, 2014.

\bibitem{Aarabi2017}
------, ``{Seizure prediction in patients with focal hippocampal epilepsy},''
  \emph{Clinical Neurophysiology}, vol. 128, no.~7, pp. 1299--1307, jul 2017.

\bibitem{LeVanQuyen2005}
M.~{Le Van Quyen} \emph{et~al.}, ``{Preictal state identification by
  synchronization changes in long-term intracranial EEG recordings},''
  \emph{Clinical Neurophysiology}, vol. 116, no.~3, pp. 559--568, 2005.

\bibitem{Mirowski2009}
P.~Mirowski \emph{et~al.}, ``{Classification of patterns of EEG synchronization
  for seizure prediction},'' \emph{Clinical Neurophysiology}, vol. 120, no.~11,
  pp. 1927--1940, 2009.

\bibitem{Stober2015}
S.~Stober \emph{et~al.}, ``{Deep Feature Learning for EEG Recordings},''
  \emph{Arxiv}, pp. 1--24, nov 2015.

\bibitem{Simard2003}
P.~Simard, D.~Steinkraus, and J.~Platt, ``{Best practices for convolutional
  neural networks applied to visual document analysis},'' in \emph{Seventh
  International Conference on Document Analysis and Recognition, 2003.
  Proceedings.}, vol.~1.\hskip 1em plus 0.5em minus 0.4em\relax IEEE Comput.
  Soc, 2003, pp. 958--963.

\bibitem{Mirowski2008a}
P.~W. Mirowski \emph{et~al.}, ``{Comparing SVM and convolutional networks for
  epileptic seizure prediction from intracranial EEG},'' in \emph{Proceedings
  of the 2008 IEEE Workshop on Machine Learning for Signal Processing, MLSP
  2008}, 2008, pp. 244--249.

\bibitem{Bandarabadi2015}
M.~Bandarabadi \emph{et~al.}, ``{On the proper selection of preictal period for
  seizure prediction},'' \emph{Epilepsy and Behavior}, vol.~46, pp. 158--166,
  2015.

\bibitem{Freestone2017}
D.~R. Freestone, P.~J. Karoly, and M.~J. Cook, ``{A forward-looking review of
  seizure prediction},'' \emph{Current Opinion in Neurology}, vol.~30, no.~2,
  pp. 167--173, 2017.

\bibitem{Osorio2011}
I.~Osorio \emph{et~al.}, \emph{{Epilepsy: The Intersection of Neuorsciences,
  Biology; Mathematics, Engineering , and Physics}}.\hskip 1em plus 0.5em minus
  0.4em\relax CRC press, 2011.

\bibitem{Goldberger2000}
A.~L. Goldberger \emph{et~al.}, ``{PhysioBank, PhysioToolkit, and PhysioNet},''
  \emph{Circulation}, vol. 101, no.~23, pp. E215--20, 2000.

\bibitem{Shoeb2009a}
A.~H. Shoeb, ``{Application of machine learning to epileptic seizure onset
  detection and treatment},'' Ph.D. dissertation, Massachusetts Institute of
  Technology, 2009.

\bibitem{Broock1996}
M.~C. Pardo and J.~A. Pardo, ``{A test for independence based on the
  Information Energy},'' \emph{Journal of the Franklin Institute}, vol. 331,
  no.~1, pp. 13--22, jan 1994.

\bibitem{kemp2003}
B.~Kemp and J.~Olivan, ``{European data format 'plus' (EDF+), an EDF alike
  standard format for the exchange of physiological data},'' \emph{Clinical
  Neurophysiology}, vol. 114, no.~9, pp. 1755--1761, sep 2003.

\bibitem{Shoeb2010}
A.~Shoeb and J.~Guttag, ``{Application of Machine Learning To Epileptic Seizure
  Detection},'' \emph{Proceedings of the 27th International Conference on
  Machine Learning (ICML-10)}, pp. 975--982, 2010.

\bibitem{Kim2000}
C.~H. Kim and R.~Aggarwal, ``{Wavelet transforms in power systems. I. General
  introduction to the wavelet transforms},'' \emph{Power Engineering Journal},
  vol.~14, no.~2, pp. 81--87, 2000.

\bibitem{Acar2007}
E.~Acar \emph{et~al.}, ``{Multiway analysis of epilepsy tensors},''
  \emph{Bioinformatics}, vol.~23, no.~13, pp. i10--i18, 2007.

\bibitem{Daubechies1990}
I.~Daubechies, ``{The wavelet transform, time-frequency localization and signal
  analysis},'' \emph{IEEE T. Inform. Theory}, no.~36, pp. 961--1005, 1990.

\bibitem{Krizhevsky2012}
A.~Krizhevsky, I.~Sutskever, and G.~E. Hinton, ``{ImageNet classification with
  deep convolutional neural networks},'' \emph{Communications of the ACM},
  vol.~60, no.~6, pp. 84--90, may 2017.

\bibitem{Srivastava2014}
N.~Srivastava \emph{et~al.}, ``{Dropout: A Simple Way to Prevent Neural
  Networks from Overfitting},'' \emph{Journal of Machine Learning Research},
  vol.~15, pp. 1929--1958, 2014.

\bibitem{Nair2010}
V.~Nair and G.~E. Hinton, ``{Rectified Linear Units Improve Restricted
  Boltzmann Machines},'' \emph{Proceedings of the 27th International Conference
  on Machine Learning}, no.~3, pp. 807--814, 2010.

\bibitem{Brodley1999}
C.~Brodley and M.~Friedl, ``{Identifying mislabeled training data},''
  \emph{Journal of Artificial Intelligence Research}, vol. 694, no.~i, pp.
  145--146, 1992.

\bibitem{Kubat1997}
M.~Kubat, R.~Holte, and S.~Matwin, ``{Learning when negative examples
  abound},'' in \emph{Proceedings of the 9th European Conference on Machine
  Learning}.\hskip 1em plus 0.5em minus 0.4em\relax Springer Berlin Heidelberg,
  1997, pp. 146--153.

\bibitem{Chawla2002}
N.~V. Chawla \emph{et~al.}, ``{SMOTE: Synthetic minority over-sampling
  technique},'' \emph{Journal of Artificial Intelligence Research}, vol.~16,
  pp. 321--357, 2002.

\bibitem{Zeiler2012}
M.~D. Zeiler, ``{ADADELTA: An Adaptive Learning Rate Method},'' \emph{arXiv},
  p.~6, 2012.

\bibitem{chollet2015}
F.~Chollet, ``keras,'' 2015.

\bibitem{TheanoDevelopmentTeam2016}
{The Theano Development Team} \emph{et~al.}, ``{Theano: A Python framework for
  fast computation of mathematical expressions},'' \emph{arXiv e-prints}, vol.
  abs/1605.0, p.~19, 2016.

\bibitem{Matthews1985a}
D.~R. Matthews, ``{Homeostasis model assessment: insulin resistance and
  beta-cell function from fasting plasma glucose and insulin concentrations in
  man},'' \emph{Diabetologia}, vol.~28, pp. 412--419, 1985.

\end{thebibliography}

\end{document}